\documentclass[fleqn,10pt]{wlscirep}
\usepackage[utf8]{inputenc}
\usepackage[T1]{fontenc}
\usepackage{graphicx}
\usepackage{booktabs}
\usepackage{array}
\usepackage{amsmath}
\usepackage{amssymb}
\usepackage{xcolor}
\usepackage{hyperref}
\usepackage{caption}
\usepackage{subcaption}
\usepackage{adjustbox}
\usepackage{float}
\usepackage{longtable}
\usepackage{placeins}
\graphicspath{{figures/}}
\hypersetup{colorlinks=true,linkcolor=black,citecolor=black,urlcolor=blue}
\newcommand{\repositoryurl}{\url{https://github.com/Dyniel/pinn-gym}}

\title{Decision-Aware Evaluation of Physics-Informed Surrogates}

\author[1,*]{Daniel Cie\'slak}
\author[1]{Andrzej Czy\.zewski}
\affil[1]{Gda\'nsk University of Technology, Gda\'nsk, Poland}
\affil[*]{Corresponding author: \href{mailto:daniel.cieslak@pg.edu.pl}{daniel.cieslak@pg.edu.pl}}


\begin{abstract}
Physics-informed machine learning is often assessed by curve error, although engineering use depends on downstream decisions: ranking candidates, avoiding infeasible designs and limiting regret. We introduce \emph{pinn-gym}, an open benchmark for material-conditioned lattice design that couples a transparent reduced-order crush-and-impact oracle with five printable polymer cards, dimensionless force-response targets and a protocol spanning curve fidelity, physical admissibility, top-$k$ retrieval and mass regret.

Across per-material, pooled and cross-material settings, low nRMSE is frequently insufficient to identify useful design selections. Physics-informed losses alter trade-offs rather than monotonically improving all metrics, and dimensionless conditioning improves comparability without making transfer symmetric. The benchmark is not a certified material model; within the released oracle, candidate generator and material cards, \emph{pinn-gym} provides a reproducible testbed for evaluating PIML surrogates as decision systems rather than curve predictors alone.
\end{abstract}

\begin{document}
\flushbottom
\maketitle
\thispagestyle{empty}

\section*{Introduction}
Physics-informed machine learning (PIML) is increasingly used to accelerate simulation and engineering design~\cite{raissi2019pinn,karniadakis2021piml,cuomo2022scientific,wang2023expert,toscano2025pinns}. Its promise is not only better interpolation, but more reliable decisions under physical constraints. In design workflows, however, the surrogate is rarely the final object of interest. It ranks candidate geometries, rejects infeasible designs and guides expensive downstream simulation or fabrication. A model can therefore be accurate as a force--displacement predictor while still being a poor design assistant.

We introduce \emph{pinn-gym}, an open benchmark for decision-aware evaluation of material-conditioned PIML surrogates in lattice energy absorption. The benchmark defines a transparent reduced-order oracle, five printable polymer material cards, candidate pools, baseline models, physics-informed losses and metrics that jointly report curve fidelity, physical admissibility, top-$k$ retrieval and mass regret. The target domain is deliberately narrow enough to be reproducible and broad enough to stress material conditioning: the same geometry can behave differently under compliant TPU and stiff carbon-fibre-reinforced PA cards~\cite{gibson1997cellular,ashby2006metalfoams,bertoldi2017metamaterials,gu2018bioinspired,liu2024lattice,zheng2023deep}.

The central modelling choice is nondimensionalisation. Following scaling-law modelling and Buckingham-$\pi$ analysis~\cite{buckingham1914pi,barenblatt1996scaling}, the surrogate predicts a dimensionless force response
\begin{equation}
    \hat f(\epsilon; g^*,m^*)=\frac{F}{\sigma_y A_{\mathrm{env}}}, \qquad \epsilon=\frac{u}{L_{\mathrm{env}}},
    \label{eq:nondim}
\end{equation}
conditioned on nondimensional geometry $g^*$ and a material vector $m^*$ encoding stiffness, strength, plateau response, failure strain, density, printability and rate-sensitivity ratios. Changing the polymer is therefore an input perturbation rather than a separate regression task. This supports three evaluations: per-material specialists, one pooled material-conditioned model and zero-shot cross-material transfer.

This framing separates three questions that are often conflated. First, does low held-out curve error imply useful design ranking? Second, do physics-informed losses improve the downstream decision or merely shift the trade-off between fit and constraint satisfaction? Third, does dimensionless material conditioning make transfer symmetric across polymer regimes? The answer to all three is non-trivial. Across the released benchmark, lower nRMSE often fails to identify the best feasible designs, additional residual terms are not uniformly beneficial, and transfer remains asymmetric despite dimensionless inputs.

The contribution is therefore a benchmark and protocol, not a certified absorber. All claims are scoped to the declared oracle, candidate generator and material cards. The purpose is to make PIML evaluation closer to how surrogates are actually used: as decision systems whose curve predictions matter only insofar as they produce feasible, low-regret choices.
\section*{Results}

\begin{figure*}[t]
\centering
\includegraphics[width=\linewidth]{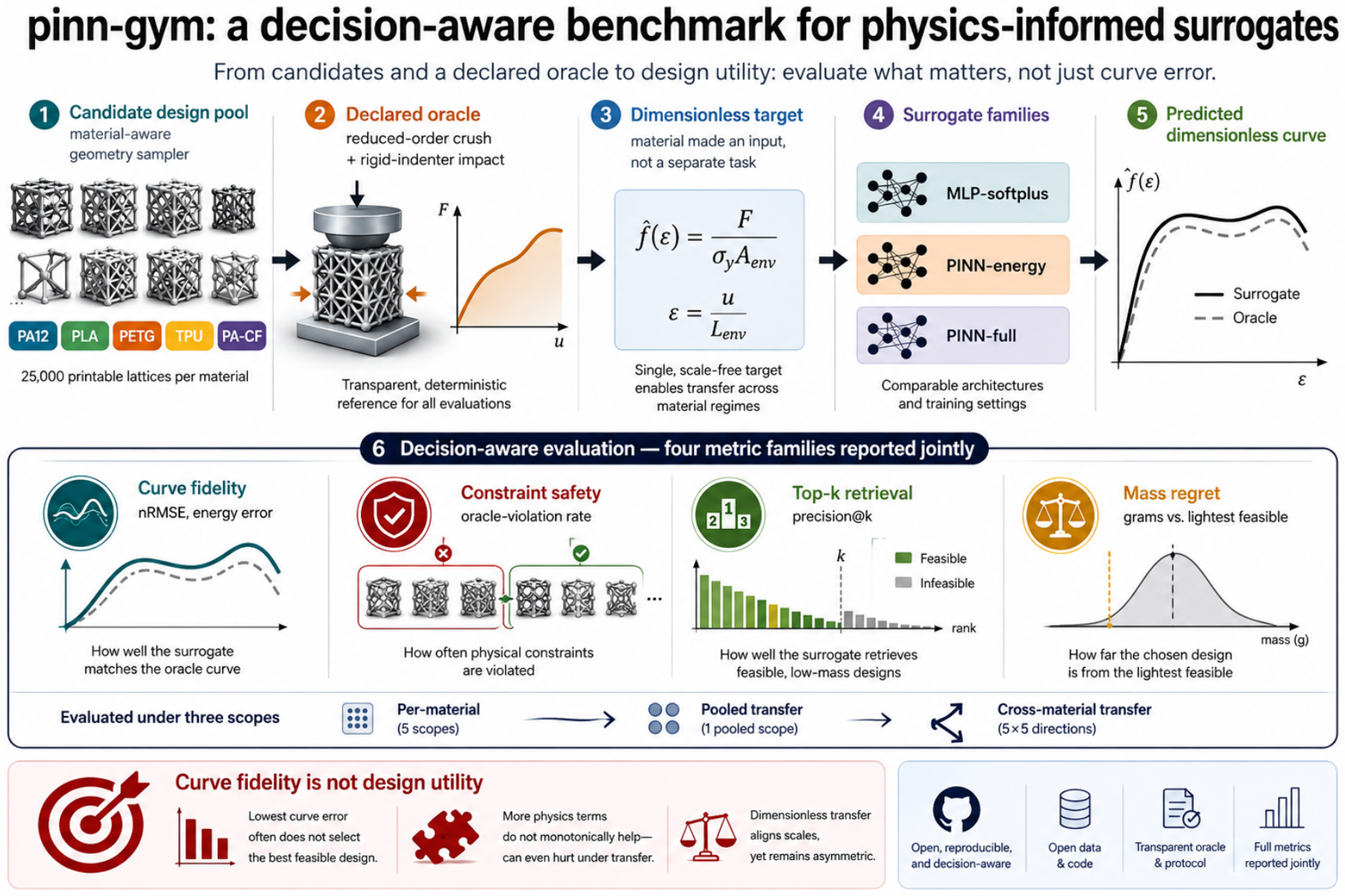}
\caption{\textbf{Overview of the \emph{pinn-gym} benchmark.} A material-aware sampler draws candidate lattice geometries for five printable polymer cards; a declared reduced-order crush-and-impact oracle labels every candidate with a force--displacement curve, absorbed energy, peak force and a feasibility outcome. Force and displacement are nondimensionalised to the target $\hat f(\epsilon)=F/(\sigma_y A_{\mathrm{env}})$, which makes the polymer an input rather than a separate regression task. Three surrogate families (data-only MLP-softplus, energy-informed PINN, composite PINN) are trained and then evaluated under three scopes (per-material, pooled, $5\times5$ transfer) through four metric families reported jointly: curve fidelity, constraint safety, top-$k$ retrieval and mass regret. The central result of the benchmark is that these families do not agree: the surrogate with the lowest curve error is frequently not the one that selects the best feasible design.}
\label{fig:overview}
\end{figure*}

\subsection*{What the benchmark measures, and how the study is organised}
\emph{pinn-gym} is a controlled diagnostic environment, not a material-property claim. The oracle is transparent, deterministic and shared across all models, so differences between methods reflect surrogate behaviour under the same labelling rule. A higher-fidelity FEM or experimental oracle would change the absolute numbers, but not the need to evaluate prediction, constraint satisfaction and selection jointly.

The study has a simple structure (Table~\ref{tab:study_structure}). We first describe the task and the feasible-set structure induced by five material cards. We then compare per-material specialists, pooled material-conditioned models and $5\times5$ zero-shot transfers. The reported metrics are deliberately heterogeneous: curve nRMSE measures interpolation, violation rate measures safety under the oracle, precision-at-$k$ measures retrieval of feasible designs, and regret-at-$k$ measures the mass penalty of the selected design. No single metric is treated as sufficient.

\begin{table}[t]
\centering
\small
\caption{Study structure and the role of each experiment block.}
\label{tab:study_structure}
\begin{tabular}{@{}p{0.25\linewidth}p{0.64\linewidth}@{}}
\toprule
Block & Purpose \\
\midrule
Task definition & Defines the oracle-labelled material-conditioned candidate pools and feasible-set structure. \\
Per-material models & Tests whether curve fidelity predicts design ranking when each material has its own specialist. \\
Pooled models & Tests whether one dimensionless material-conditioned surrogate can replace specialists. \\
Transfer matrix & Tests whether a checkpoint trained on one material transfers symmetrically to another. \\
Seed repeats & Tests whether decision-aware conclusions survive stochastic retraining. \\
\bottomrule
\end{tabular}
\end{table}

\subsection*{pinn-gym defines a material-conditioned decision benchmark}
The benchmark holds out oracle-labelled candidate pools for five material cards and evaluates all surrogates on identical inputs. A material-aware sampler, described in Methods, centres each pool around the declared peak-force frontier so that easy and hard candidates coexist wherever the fixture constraints admit feasible designs. The fixture constraints are fixed across materials at a peak-force limit of $3500$~N and a minimum crush distance of $40$~mm under a $6$~kg, $0.5$~m drop ($29.43$~J). Differences in feasibility therefore come from the material card and candidate geometry, not from re-tuning the task per polymer.

\begin{figure}[t]
\centering
\includegraphics[width=\linewidth]{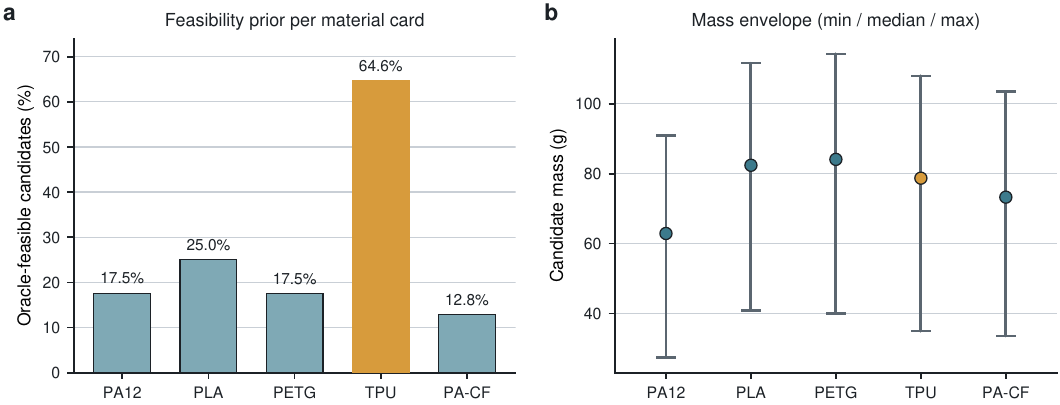}
\caption{\textbf{One oracle, five design regimes.} (\textbf{a}) Fraction of oracle-feasible candidates in each held-out pool under the fixed fixture; the compliant TPU card (highlighted) is feasible for $64.6\%$ of candidates, whereas the stiff PA-CF card is feasible for only $12.8\%$. (\textbf{b}) The corresponding candidate-mass envelope (minimum, median and maximum) shows that the cards also occupy different regions of the mass axis. Because feasibility differs by a factor of five across cards, feasibility priors are treated as part of the benchmark rather than as incidental dataset statistics; they set the ceiling on what any surrogate can retrieve.}
\label{fig:eval_feasibility_current}
\end{figure}

The first benchmark result is descriptive but important: the same oracle creates very different design regimes across material cards (Table~\ref{tab:eval_pools}, Fig.~\ref{fig:eval_feasibility_current}). Compliant and stiff polymers occupy different regions of mass, peak force and crush distance even under the same geometric envelope. This matters for PIML evaluation because a single curve-error number does not reveal whether a target pool contains many feasible low-mass designs, very few feasible designs or no feasible designs under the declared fixture. We therefore treat feasibility priors as part of the benchmark, not as incidental dataset statistics.

\begin{table}[t]
\centering
\caption{Held-out evaluation pools by material card. Feasibility is assigned by the declared numerical oracle. Mass values are reported in grams.}
\label{tab:eval_pools}
\begin{tabular}{lrrrrr}
\toprule
Material & Candidates & Feasible & Feasible [\%] & Median mass & Mass range \\
\midrule
PA12 & 25,000 & 4,383 & 17.53 & 62.9 & 27.4--90.9 \\
PLA & 25,000 & 6,254 & 25.02 & 82.4 & 40.8--111.6 \\
PETG & 25,000 & 4,371 & 17.48 & 84.1 & 40.0--114.3 \\
TPU & 25,000 & 16,149 & 64.60 & 78.7 & 35.0--108.0 \\
PA-CF & 25,000 & 3,211 & 12.84 & 73.3 & 33.6--103.5 \\
\bottomrule
\end{tabular}
\end{table}

\subsection*{Lower curve error does not imply better design rankings}
\begin{figure*}[t]
\centering
\includegraphics[width=\linewidth]{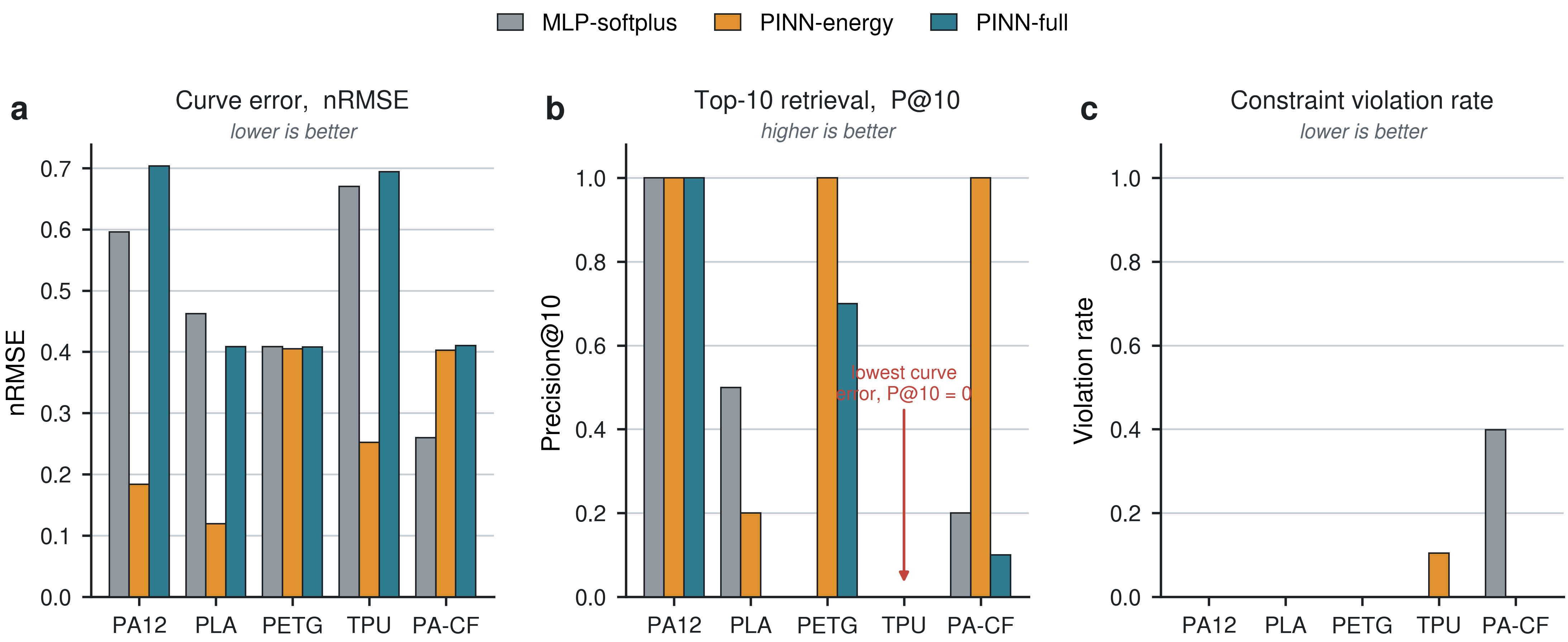}
\caption{\textbf{Curve accuracy and design utility disagree within a single material card.} Per-material surrogate metrics for the three model families across (\textbf{a}) curve fidelity (nRMSE, lower is better), (\textbf{b}) top-$10$ feasible retrieval (P@$10$, higher is better) and (\textbf{c}) oracle-violation rate (lower is better). The highlighted case marks a card on which the lowest-curve-error model retrieves no feasible design in its top ten. Read together with Table~\ref{tab:per_material}, the panels show that the best curve predictor is frequently neither the safest nor the most useful ranker.}
\label{fig:per_material_metrics}
\end{figure*}

The clearest evidence for reporting multiple metric families is that, on a single card, curve accuracy and design ranking can disagree (Table~\ref{tab:per_material}, Fig.~\ref{fig:per_material_metrics}). On PLA, the energy-informed PINN attains the lowest curve error in the main protocol (nRMSE $0.119$), but the data-only MLP retrieves more oracle-feasible candidates in the top-10 screening set despite a higher curve error (nRMSE $0.462$). Because P@$10$ is a deliberately small-budget screening metric, these values should be read as evidence of metric disagreement rather than as a precise estimate of a stable performance ratio. On PA-CF the reversal is sharper: the lowest-curve-error model (MLP, nRMSE $0.260$) endorses unsafe designs at a $0.40$ oracle-violation rate and reaches only $P@10=0.2$, whereas the higher-error energy PINN (nRMSE $0.402$) is constraint-safe and retrieves every feasible design in its top ten ($P@10=1.0$). A curve-error leaderboard would therefore miss decision-relevant failures that become visible only when curve fidelity, feasibility and retrieval are reported jointly.

This is the central scientific message of the benchmark. For design, the relevant question is not simply whether the surrogate predicts $F(u)$ accurately, but whether its prediction induces a useful ordering over candidate geometries. Precision-at-$k$ and regret-at-$k$ expose this ordering. We use P@$10$ as an interpretable small-budget screening metric, not as a continuous estimator of ranking quality; broader values of $k$ are included in the evaluation harness to check whether the same disagreement persists beyond the top-ten budget. Regret is especially useful because it converts ranking quality into an engineering quantity: the mass penalty paid relative to the lightest oracle-feasible candidate available in the pool. When the feasible set is empty under the declared fixture, ranking metrics are reported as non-applicable rather than as failures. This distinction is essential for stiff cards and prevents the benchmark from confusing sampler and constraint geometry with model quality.

\begin{table*}[t]
\centering
\small
\caption{\textbf{Material-specific surrogate evaluation (main protocol).} Specialists trained and evaluated on a single card, on the $25{,}000$-candidate held-out pools; rows are ordered by curve error within each card to make the accuracy--utility dissociation visible. Lower nRMSE, violation and regret are better; higher P@10 is better. $\infty$ marks a non-empty feasible pool from which the model retrieved no feasible design in its top ten.}
\label{tab:per_material}
\begin{tabular}{llrrrr}
\toprule
Material & Method & nRMSE $\downarrow$ & Violation $\downarrow$ & P@10 $\uparrow$ & Regret@10 [g] $\downarrow$ \\
\midrule
PA12 & PINN-energy & 0.184 & 0.000 & 1.000 & 0.28 \\
PA12 & MLP-softplus & 0.596 & 0.000 & 1.000 & 1.03 \\
PA12 & PINN-full & 0.704 & 0.000 & 1.000 & 0.57 \\
\addlinespace
PLA & PINN-energy & 0.119 & 0.000 & 0.200 & 5.00 \\
PLA & PINN-full & 0.408 & 0.000 & 0.000 & $\infty$ \\
PLA & MLP-softplus & 0.462 & 0.000 & 0.500 & 5.82 \\
\addlinespace
PETG & PINN-energy & 0.405 & 0.000 & 1.000 & 5.92 \\
PETG & PINN-full & 0.408 & 0.000 & 0.700 & 5.92 \\
PETG & MLP-softplus & 0.408 & 0.000 & 0.000 & $\infty$ \\
\addlinespace
TPU & PINN-energy & 0.252 & 0.104 & 0.000 & $\infty$ \\
TPU & MLP-softplus & 0.670 & 0.000 & 0.000 & $\infty$ \\
TPU & PINN-full & 0.694 & 0.000 & 0.000 & $\infty$ \\
\addlinespace
PA-CF & MLP-softplus & 0.260 & 0.399 & 0.200 & 0.00 \\
PA-CF & PINN-energy & 0.402 & 0.000 & 1.000 & 6.89 \\
PA-CF & PINN-full & 0.410 & 0.000 & 0.100 & 0.00 \\
\bottomrule
\end{tabular}
\end{table*}

\subsection*{Physics-informed losses change the trade-off, not just the fit}
The energy-informed model and the composite PINN should not be read as a monotonic ladder in which more residual terms automatically imply a better surrogate. The energy residual often acts as a useful prior over the feasible region: it shapes the predicted curve in a way that can reduce physically inadmissible selections and improve energy consistency. The additional peak-force, monotonicity-after-densification and smoothness penalties are individually defensible, but their scalarised combination changes the optimisation landscape. In several settings the composite objective improves one quantity while weakening another.

This behaviour is not a reason to discard physics-informed learning. It is the reason to benchmark it decision-wise. In \emph{pinn-gym}, the correct question is not ``does the constraint help?'' but ``which part of the design objective does the constraint help?'' A residual can improve energy-integral behaviour, reduce violation rate or regularise extrapolation without being the model with the lowest nRMSE. The loss-weight ablation in Table~\ref{tab:loss_weight_ablation_zip} makes this dependence explicit. A boundary-only objective collapses to high predicted-feasible rates and poor curve fit, while energy and composite objectives reduce curve error by an order of magnitude. Strengthening different physics terms changes the trade-off surface: the peak-strong setting gives the lowest macro nRMSE in this ablation ($0.219$) and zero mean violation, whereas the default composite gives slightly higher nRMSE ($0.250$) but a different feasible-retrieval profile. This is exactly the behaviour the benchmark is designed to expose.

\begin{table*}[t]
\centering
\small
\caption{\textbf{Physics-loss weight ablation (revision protocol).} Effect of the physics-loss weights on pooled \texttt{pinn\_full}. Values are macro-averaged over the five material cards; regret@10 is averaged only over cards on which a finite regret is defined. Lower nRMSE, energy error, violation and regret are better; higher P@10 is better. The default composite row uses the declared weights of Table~\ref{tab:loss_weights}. Because this ablation is run on the re-sampled revision pools, its absolute numbers are a within-protocol sensitivity analysis and are not directly comparable to the main 25{,}000-candidate pools.}
\label{tab:loss_weight_ablation_zip}
\begin{tabular}{@{}lrrrrrrrrrrr@{}}
\toprule
& \multicolumn{5}{c}{Loss weights} & \multicolumn{6}{c}{Macro-averaged metrics} \\
\cmidrule(lr){2-6}\cmidrule(lr){7-12}
Ablation & $w_b$ & $w_E$ & $w_p$ & $w_m$ & $w_s$ & nRMSE $\downarrow$ & Energy err.\ [J] $\downarrow$ & Violation $\downarrow$ & Pred.-feas. & P@10 $\uparrow$ & Regret@10 [g] $\downarrow$ \\
\midrule
Boundary only & 0.05 & 0 & 0 & 0 & 0 & 1.645 & 35.496 & 0.692 & 0.800 & 0.060 & 2.691 \\
Energy only & 0.05 & 0.20 & 0 & 0 & 0 & 0.286 & 10.431 & 0.440 & 0.007 & 0.300 & 2.180 \\
Default composite & 0.05 & 0.20 & 0.10 & 0.05 & 0.02 & 0.250 & 10.078 & 0.636 & 0.007 & 0.320 & 2.180 \\
Energy-strong & 0.05 & 0.50 & 0.10 & 0.05 & 0.02 & 0.272 & 13.708 & 0.200 & 0.001 & 0.380 & 1.939 \\
Peak-strong & 0.05 & 0.20 & 0.30 & 0.05 & 0.02 & 0.219 & 9.733 & 0.000 & 0.000 & 0.380 & 1.939 \\
\bottomrule
\end{tabular}
\end{table*}

\begin{table}[t]
\centering
\small
\caption{Declared default physics-loss weights used for the composite PINN. The weights are treated as benchmark choices and ablated rather than as universal constants.}
\label{tab:loss_weights}
\begin{tabular}{@{}lc@{}}
\toprule
Loss term & Weight \\
\midrule
Boundary consistency & 0.05 \\
Energy residual & 0.20 \\
Peak-force penalty & 0.10 \\
Post-densification monotonicity & 0.05 \\
Curvature smoothness & 0.02 \\
\bottomrule
\end{tabular}
\end{table}

\subsection*{Pooled material conditioning fits curves but does not unify decisions}
\begin{figure}[t]
\centering
\includegraphics[width=\linewidth]{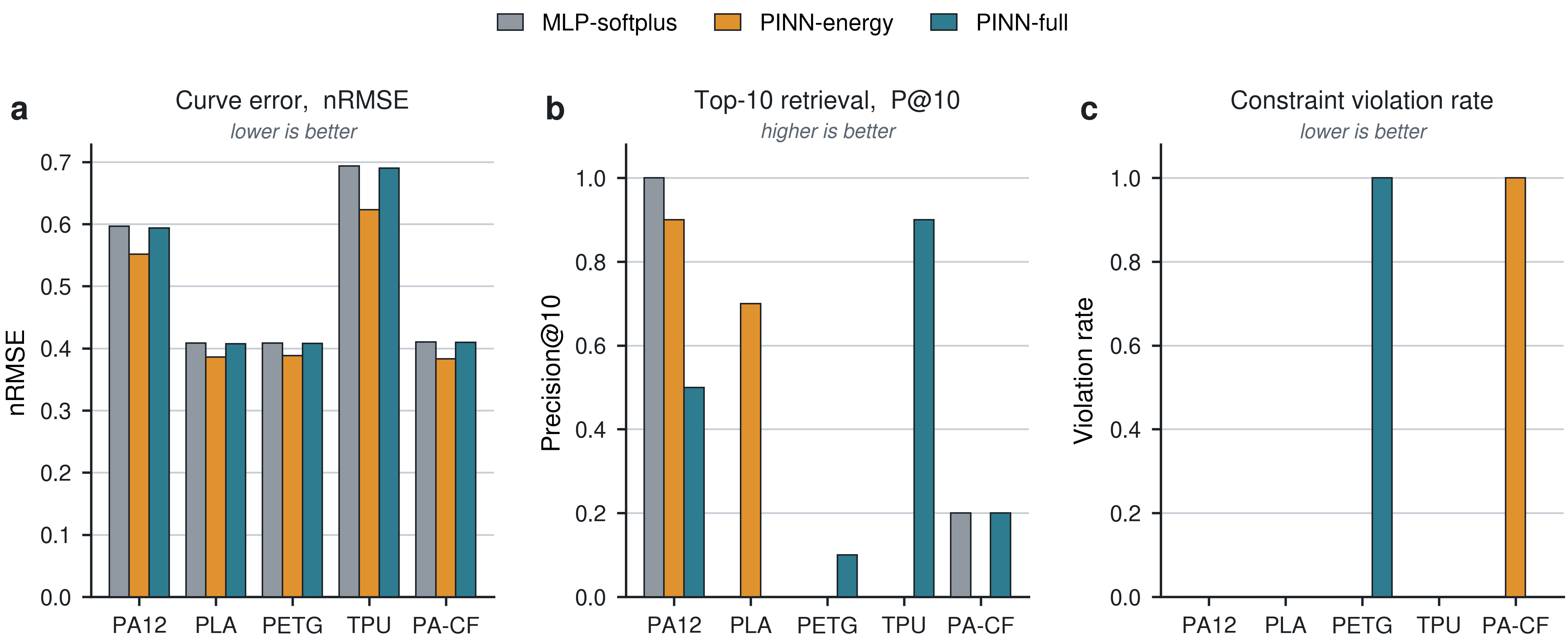}
\caption{\textbf{Pooled material-conditioned evaluation.} A single nondimensional surrogate can fit multiple material cards, but curve error, oracle violations and top-$k$ retrieval remain partially decoupled.}
\label{fig:pooled_metrics}
\end{figure}

The pooled setting tests whether one nondimensional, material-conditioned surrogate can replace five specialists. On the main held-out pools, a single pooled network learns the material conditioning and reaches moderate curve error on every card, with nRMSE between $0.38$ and $0.69$. However, the decision metrics remain target-specific. On TPU, the composite PINN reaches $P@10=0.9$, whereas the energy-informed PINN trained on the same pooled data retrieves no oracle-feasible design in its top ten ($P@10=0$). On PETG and PA-CF, some pooled variants select predicted-feasible candidates that the oracle rejects entirely, giving violation rates of $1.0$. Thus pooled curve accuracy is necessary but not sufficient: one model can fit all five material cards while still making poor design selections on individual targets. The per-card numbers are reported in Table~\ref{tab:pooled} and Fig.~\ref{fig:pooled_metrics}.

\begin{table*}[t]
\centering
\small
\caption{\textbf{Pooled material-conditioned surrogate evaluation (main protocol).} A single network is trained across all five cards and evaluated per card on the $25{,}000$-candidate held-out pools; these are the numbers visualised in Fig.~\ref{fig:pooled_metrics}. A pooled model can reach moderate curve error on every card while its decision metrics remain strongly card-specific. Lower nRMSE, violation and regret are better; higher P@10 is better.}
\label{tab:pooled}
\begin{tabular}{llrrrr}
\toprule
Material & Method & nRMSE $\downarrow$ & Violation $\downarrow$ & P@10 $\uparrow$ & Regret@10 [g] $\downarrow$ \\
\midrule
PA12 & MLP-softplus & 0.597 & 0.000 & 1.000 & 0.00 \\
PA12 & PINN-energy & 0.551 & 0.000 & 0.900 & 1.68 \\
PA12 & PINN-full & 0.594 & 0.000 & 0.500 & 16.07 \\
\addlinespace
PLA & MLP-softplus & 0.408 & 0.000 & 0.000 & $\infty$ \\
PLA & PINN-energy & 0.386 & 0.000 & 0.700 & 8.31 \\
PLA & PINN-full & 0.407 & 0.000 & 0.000 & $\infty$ \\
\addlinespace
PETG & MLP-softplus & 0.409 & 0.000 & 0.000 & $\infty$ \\
PETG & PINN-energy & 0.388 & 0.000 & 0.000 & $\infty$ \\
PETG & PINN-full & 0.408 & 1.000 & 0.100 & 26.07 \\
\addlinespace
TPU & MLP-softplus & 0.694 & 0.000 & 0.000 & $\infty$ \\
TPU & PINN-energy & 0.623 & 0.000 & 0.000 & $\infty$ \\
TPU & PINN-full & 0.690 & 0.000 & 0.900 & 12.71 \\
\addlinespace
PA-CF & MLP-softplus & 0.410 & 0.000 & 0.200 & 0.00 \\
PA-CF & PINN-energy & 0.383 & 1.000 & 0.000 & $\infty$ \\
PA-CF & PINN-full & 0.409 & 0.000 & 0.200 & 14.83 \\
\bottomrule
\end{tabular}
\par\vspace{3pt}{\footnotesize\raggedright \textit{Columns.} nRMSE: curve fidelity (lower is better); Violation: fraction of predicted-feasible candidates the oracle rejects; P@10: top-10 feasible-retrieval precision; Regret@10: mass gap in grams to the lightest oracle-feasible design. \textit{Symbols.} $\infty$ marks a non-empty feasible pool from which the model retrieved no feasible candidate in its top ten (a decision failure, not a missing pool); N/A would mark a pool with no oracle-feasible candidate, which does not occur in the main 25{,}000-candidate pools. All rows use the main protocol.\par}
\end{table*}

We repeated the pooled experiment after correcting pooled step scaling and tuning capacity. This second protocol confirms the same message in a less pessimistic setting: tuned pooled models can reach competitive curve errors and sometimes strong decision performance, but the best trade-off is still material-dependent. In the capacity ablation, increasing width and depth improves curve accuracy, whereas decision metrics saturate or move non-monotonically. In the loss-weight ablation, energy, peak-force and composite terms improve different parts of the objective rather than forming a universal ordering.

The practical conclusion is not that pooled conditioning fails. It is that pooled PIML must be evaluated per target material and per downstream use. For early screening, a slightly less accurate model with lower violation rate may be preferable. For curve interpolation, the opposite may hold. Reporting only a pooled average nRMSE would hide this distinction.

\subsection*{Dimensionless transfer aligns scales but stays asymmetric}
\begin{figure*}[t]
\centering
\includegraphics[width=\linewidth]{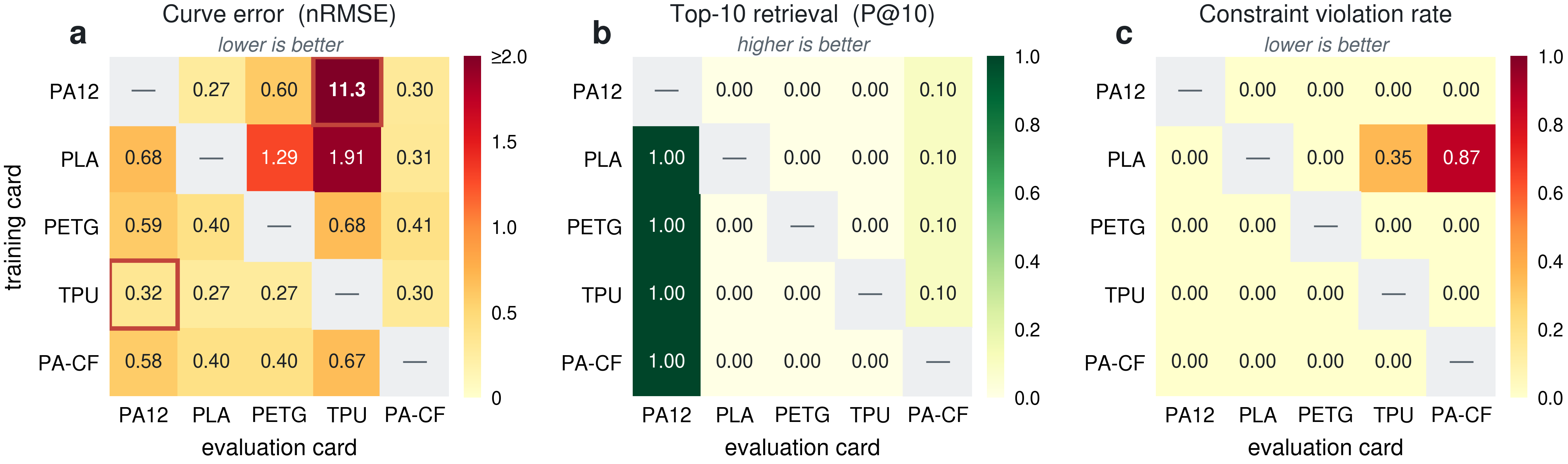}
\caption{\textbf{Dimensionless conditioning aligns scales but leaves cross-material transfer strongly asymmetric.} Cross-material transfer of the PINN-energy model across five material cards. Rows denote the material used for training and columns the material used for evaluation; diagonal entries are omitted because self-material performance is reported separately. Panel A shows curve-prediction error as nRMSE (highlighted cells mark the PA12--TPU pair, where the two directions differ by more than thirtyfold). Panel B reports top-10 retrieval precision for oracle-feasible candidates under the predicted design score. Panel C reports the fraction of predicted-feasible candidates that violate oracle constraints. Curve-level transfer and design-selection utility are not equivalent, and neither is symmetric across training and evaluation cards.}
\label{fig:transfer_pinn_energy}
\end{figure*}

The $5\times 5$ transfer matrix asks a different question from pooled training. Each off-diagonal entry is produced by taking a checkpoint trained on one material, re-scaling its inputs with the target material's nondimensional groups and evaluating it on the target pool without retraining or fine-tuning (Fig.~\ref{fig:transfer_pinn_energy}). If nondimensionalisation removed all material-regime bias, the matrix would be close to symmetric. It is not. The clearest case is the PA12--TPU pair: a TPU-trained checkpoint transfers onto the PA12 pool with curve error $\mathrm{nRMSE}=0.32$ and recovers the entire feasible top-10 ($P@10=1.0$), whereas the reverse direction, PA12 onto TPU, degrades to $\mathrm{nRMSE}=11.3$ and retrieves nothing ($P@10=0$)---a more than thirty-fold gap in curve error between two directions that the dimensionless formulation makes nominally equivalent. The asymmetry is not confined to that pair: the PLA--PA12 direction shows the same ordering on a smaller scale ($0.27$ versus $0.68$), and several transfers preserve acceptable curve error while still retrieving no feasible top-ranked candidate. Transfers from compliant regimes onto stiffer ones are systematically easier than the reverse.

This result should be read constructively. The transfer matrix is a diagnostic for future PIML methods rather than a statement that transfer is impossible. It identifies where a single-material checkpoint has learned response-shape statistics that do not extrapolate across material groups, and it separates three kinds of transfer success: curve prediction, feasible retrieval and constraint safety. A new surrogate, neural operator or meta-learning method can be evaluated against exactly the same matrix and can claim improvement only when it reduces this asymmetry in the relevant metric family rather than in one curve number.

\subsection*{Repeated seeds support the benchmark message}
To reduce the risk that the conclusions are single-run artefacts, the revision experiments repeat the PINN-energy and PINN-full variants over five seeds and summarise mean and standard deviation per card (Table~\ref{tab:seed_repeats_zip}). These repeats are used to test the qualitative benchmark message, not to turn the single-seed main protocol into a definitive ranking of model families. The repeated runs preserve the main conclusion that physics-informed losses alter the balance between curve error, predicted feasibility, violation rate and regret, and that these outputs cannot be inferred from nRMSE alone. They also show why decision metrics should be reported with uncertainty: for TPU, PINN-full has a high nRMSE standard deviation ($0.339$) and a predicted-feasible-rate standard deviation of $0.422$, whereas other cards are more stable.

\begin{table*}[t]
\centering
\small
\caption{Five-seed repeat summary from the revision experiments. Values are mean $\pm$ standard deviation by material and method. Regret@10 is reported according to the revision protocol and is finite only where a feasible selected design exists.}
\label{tab:seed_repeats_zip}
\begin{tabular}{@{}llcccc@{}}
\toprule
Material & Method & nRMSE & Pred.-feas. & Violation & Regret@10 [g] \\
\midrule
PA12 & PINN-energy & $0.196\pm0.065$ & $0.001\pm0.001$ & $0.000\pm0.000$ & $0.000\pm0.000$ \\
PA12 & PINN-full & $0.388\pm0.174$ & $0.011\pm0.024$ & $0.034\pm0.076$ & $0.000\pm0.000$ \\
PA-CF & PINN-energy & $0.108\pm0.016$ & $0.000\pm0.000$ & $0.000\pm0.000$ & $0.453\pm0.754$ \\
PA-CF & PINN-full & $0.241\pm0.152$ & $0.008\pm0.017$ & $0.027\pm0.060$ & $0.755\pm1.557$ \\
PETG & PINN-energy & $0.111\pm0.010$ & $0.000\pm0.000$ & $0.200\pm0.447$ & $1.751\pm2.643$ \\
PETG & PINN-full & $0.202\pm0.094$ & $0.006\pm0.014$ & $0.097\pm0.218$ & $0.574\pm1.230$ \\
PLA & PINN-energy & $0.094\pm0.010$ & $0.000\pm0.000$ & $0.000\pm0.000$ & $0.582\pm0.797$ \\
PLA & PINN-full & $0.205\pm0.124$ & $0.008\pm0.018$ & $0.204\pm0.445$ & $0.290\pm0.648$ \\
TPU & PINN-energy & $0.195\pm0.061$ & $0.001\pm0.001$ & $0.000\pm0.000$ & $11.663\pm3.323$ \\
TPU & PINN-full & $0.413\pm0.339$ & $0.444\pm0.422$ & $0.145\pm0.178$ & $5.609\pm4.708$ \\
\bottomrule
\end{tabular}
\end{table*}

\section*{Discussion}
\emph{pinn-gym} makes physics-informed learning measurable at the point where it matters for engineering design: candidate selection. The benchmark does not ask only whether a surrogate predicts a force--displacement curve. It asks whether that prediction leads to physically admissible, low-mass and high-value choices under a declared oracle.

\paragraph*{Decision-aware evaluation is the main contribution.}
Curve fidelity, physical admissibility and design utility are not interchangeable. In the benchmark, a surrogate can improve nRMSE while worsening top-$k$ retrieval, or reduce oracle violations while sacrificing pointwise fit. This is expected when a regression model is used as a ranking function. Engineering PIML papers should therefore report prediction metrics, physical metrics and decision metrics together.

\paragraph*{Physics-informed losses are design priors.}
The energy residual is useful because it shapes the feasible decision boundary, not because it universally minimises every error measure. Additional peak-force, monotonicity and smoothness penalties are physically reasonable, but scalarising several residuals can move the optimiser away from the best trade-off for a given design workflow~\cite{wang2021ntk,wang2022respecting,krishnapriyan2021characterizing,bischof2021multiobjective,daw2023mitigating}. The practical recommendation is to add physics terms one at a time, declare their weights and report effects on curve error, violation rate and regret.

\paragraph*{Dimensionless conditioning helps but does not solve transfer.}
The pooled experiments show the positive result: one nondimensional, material-conditioned network can cover several polymer cards. The transfer matrix gives the caution: changing material cards is not a symmetric perturbation. A checkpoint trained in one regime can carry response-shape biases that fail in another. This makes transfer matrices a useful diagnostic for neural operators, uncertainty-aware surrogates and active-learning policies, not merely for the baselines used here.

\paragraph*{The benchmark is intentionally oracle-scoped.}
The released oracle is not a substitute for FEM or experiment. Its value is that it is transparent, cheap and fixed, which makes it useful for method comparison. A future high-fidelity version could replace the oracle while preserving the same protocol: evaluate curve prediction, physical admissibility, retrieval and regret jointly. Whenever aggregate regret is reported, it is computed only over finite selected designs and should be interpreted together with the corresponding retrieval-failure rate. Infinite regret is therefore a categorical decision failure in the tables, not a numeric value averaged with finite masses.
\section*{Limitations}
The benchmark is oracle-scoped by design. The five material cards encode declared assumptions about density, stiffness, yield, plateau response, failure strain, printable feature limits and strain-rate sensitivity. They are not calibrated to a specific filament batch, printer, build orientation or post-processing condition. The reduced-order crush model and rigid-indenter impact integration provide a transparent labelling rule rather than certified finite-element predictions. The fixture limits (peak force $3500$~N, crush distance $40$~mm and $29.43$~J impact energy) define the benchmark task, not an experimentally certified safety margin.

The ranking metrics must also be interpreted with the feasible set in mind, and the tables separate two cases that are easy to conflate. When the declared candidate pool for a material contains no oracle-feasible design under the fixed fixture, precision-at-$k$ and regret-at-$k$ are non-applicable (denoted \textbf{N/A}) rather than evidence that a model failed to select a feasible design. When a feasible set exists but a model retrieves none of its members in the top-$k$, regret is infinite (denoted \textbf{$\infty$}). These cases are kept distinct in every table because conflating them would overstate model failure and understate the role of candidate-pool geometry.

Finally, the pooled-tuning results are intended as a sensitivity analysis of the benchmark and modelling pathway, not as a claim that the selected architecture is globally optimal. The benchmark should therefore be read as a controlled, reproducible environment for comparing PIML design surrogates under a declared oracle. It is complementary to, not a substitute for, high-fidelity FEM and physical testing campaigns required before any lattice absorber can be deployed.

\section*{Methods}
The benchmark, surrogate code, oracle definition, candidate pools, training scripts and evaluation harness are available through the project repository at \repositoryurl{}. Run metadata for every reported number, including the random seed or seed list, checkpoint hashes and candidate-pool generator state, is recorded in the per-run JSON files distributed with the archive so that every figure and table in the manuscript can be regenerated end to end.

\subsection*{Main and revision experiment protocols}
The manuscript combines two compatible evaluation protocols. The main benchmark protocol uses the 25,000-candidate held-out pools per material to compare per-material specialists, pooled models and cross-material transfer under a single declared oracle. The revision protocol uses the same oracle, material cards, metrics and reporting conventions, but adds pooled-step-scaling correction, physics-loss weight ablation, pooled capacity tuning and five-seed repeats. We report these revision experiments separately when their candidate-pool geometry differs from the main 25,000-candidate pools; in particular, top-$k$ retrieval and regret are treated as non-applicable when a revision pool has no oracle-feasible candidate under the fixed fixture. This separation prevents model failure from being conflated with absence of feasible designs in the candidate pool.

\subsection*{Declared numerical oracle}
The declared oracle combines a reduced-order layered crush model with a rigid-indenter dynamic impact integrator. Each candidate is first mapped to geometric descriptors (cell size, wall and cap thicknesses, edge-rib geometry, envelope dimensions, derived relative density), and then to a predicted mass, a force--displacement curve $F(u)$, an absorbed-energy integral $E_{\mathrm{abs}}=\int F\,du$, a peak force, and an impact-survival outcome from a $6$~kg, $0.5$~m drop ($29.43$~J of incident kinetic energy). A candidate is labelled oracle-feasible when both the crush model and the impact integration satisfy the fixture limits $F_{\max}\leq 3500$~N and crush displacement $\geq 40$~mm. The oracle is deterministic given the geometry and the material card; it is fully described in code and contains no machine-learned components, so it can be re-implemented or audited independently of any specific simulator. The fixture limits define the benchmark task, not an experimentally certified safety margin.

\subsection*{Material cards}
Five material cards span the printable-polymer regime: PA12, PLA, PETG, TPU (thermoplastic polyurethane) and PA-CF (carbon-fibre-reinforced polyamide). Each card declares density, elastic modulus, yield strength, plateau-to-yield ratio, failure strain, $z$-direction anisotropy, printer tolerance, minimum printable feature size and a strain-rate sensitivity factor. The cards are numerical modelling inputs, not experimentally calibrated filament batches; full numerical values are tabulated in the repository so that researchers extending the benchmark can add new cards on the same footing.

\subsection*{Material-aware candidate sampling}
For each card, the sampler centres the candidate pool on the declared peak-force frontier. The relative load-bearing area scale is
\begin{equation}
    \phi_{\mathrm{peak}}=\frac{F_{\mathrm{lim}}}{\sigma_y A_{\mathrm{env}}},
\end{equation}
with $F_{\mathrm{lim}}=3500$~N. The sampling band width is clamped to $[0.10,0.22]$, and geometric bounds for walls, cells, caps, edge ribs and minimum features are adjusted using the card's minimum printable feature size and printer tolerance. The intent is to produce per-card pools that straddle the feasibility frontier where one exists inside the printable-geometry envelope: rejection-only sampling would over-represent infeasible designs on PA-CF and under-represent the easy-feasible bulk on TPU.

\subsection*{Dimensionless scaling}
For each material card, force, energy and displacement are nondimensionalised as
\begin{equation}
    F_{\mathrm{scale}}=\sigma_y A_{\mathrm{env}}, \quad E_{\mathrm{scale}}=F_{\mathrm{scale}}\,\tfrac{L_{\mathrm{env}}}{1000}, \quad \epsilon=\frac{u}{L_{\mathrm{env}}}, \quad \hat f=\frac{F}{F_{\mathrm{scale}}},
\end{equation}
with $\sigma_y$ in MPa and $A_{\mathrm{env}}$ in mm$^2$ (so $1$~MPa $=1$~N/mm$^2$). The material descriptors are nondimensionalised against PA12 as a reference card. The 10-dimensional material vector $m^*$ contains: stiffness ratio, yield ratio, plateau-to-yield ratio, failure-strain ratio, density ratio, minimum-feature ratio, tolerance ratio, $z$-anisotropy, strain-rate factor and yield-to-modulus ratio. This vector is the only material-specific input the network sees; the geometric descriptors $g^*$ are themselves ratios (relative density, cell-to-envelope, wall-to-cell, etc.) and the strain $\epsilon$ is dimensionless by construction.

\subsection*{Surrogate architecture and training}
The surrogate is a fully connected residual MLP with SiLU activations and a softplus output head that enforces non-negativity of $\hat f$. The input is the concatenation of the geometric descriptors $g^*$, the 10-dimensional material vector $m^*$ and the strain $\epsilon$. The output is the nondimensional force $\hat f(\epsilon; g^*, m^*)$ evaluated on a fixed strain grid; the full $F(u)$ curve is reconstructed by multiplying by $F_{\mathrm{scale}}$. The network has four residual blocks of width 256, layer normalisation and dropout disabled in the reported runs to keep the loss-term ablation interpretable. Three loss variants are trained from the same initialisation and the same data: \textbf{MLP-softplus} (data loss and a weak zero-displacement boundary penalty); \textbf{PINN-energy} (adds a dimensionless trapezoidal energy-integral residual); \textbf{PINN-full} (adds peak-force, post-densification monotonicity and curvature-smoothness terms).

The trained objective is
\begin{equation}
\mathcal{L}=\mathcal{L}_{\mathrm{data}}+\lambda_b\mathcal{L}_{\mathrm{boundary}}+\lambda_E\mathcal{L}_{\mathrm{energy}}+\lambda_p\mathcal{L}_{\mathrm{peak}}+\lambda_m\mathcal{L}_{\mathrm{mono}}+\lambda_s\mathcal{L}_{\mathrm{smooth}},
\end{equation}
with unused terms set to zero in each variant. The boundary term enforces $\hat f(0)=0$. The energy term matches the dimensionless trapezoidal integral $\sum_i \hat f_i\,\Delta\epsilon$ to the oracle absorbed energy divided by $E_{\mathrm{scale}}$. The peak term penalises predicted forces above a soft bound expressed in $\hat f$. The monotonicity term penalises negative derivatives after the densification strain estimated from the predicted curve. The smoothness term penalises the second-order finite difference of $\hat f$ along the strain grid. Optimisation uses Adam with an initial learning rate of $10^{-3}$, cosine decay to $10^{-5}$, batch size $512$, $300$ epochs per checkpoint and gradient clipping at $\ell_2$ norm $1.0$. The main-protocol runs use the fixed random seed \texttt{20260519}; the revision repeat experiments use five seeds as reported in Table~\ref{tab:seed_repeats_zip}; per-material specialists are trained on $50{,}000$ oracle-labelled candidates per card not included in the held-out pool, and the pooled model is trained on the union, with material card balancing per batch.

\subsection*{Evaluation metrics}
Evaluation is performed on $25{,}000$ oracle-labelled held-out candidates per material card. Metrics are stratified per card and reported jointly. Curve RMSE is in newtons, curve nRMSE divides by the per-card oracle force range, mean absolute energy-integral error is reported in joules. The decision-level metrics are: predicted-feasible rate (the fraction of candidates the surrogate would endorse as feasible at its internal threshold), physical-violation rate (the fraction of those endorsements rejected by the oracle), precision-at-$k$ for $k\in\{1,5,10,25,50\}$ on the top-ranked predicted-feasible candidates ranked by mass, and regret-at-$k$ defined as the mass gap between the lightest feasible candidate retrieved in the top-$k$ and the lightest oracle-feasible candidate in the full pool. Regret is reported in grams. We use two distinct symbols for non-finite regret and apply them consistently in every table. \textbf{N/A} marks a material pool that contains \emph{no} oracle-feasible candidate under the fixed fixture, so precision-at-$k$ and regret-at-$k$ are not applicable and must not be read as a model failure. \textbf{$\infty$} marks the opposite case: the pool does contain oracle-feasible candidates, but the model retrieved none of them in its top-$k$, so the mass penalty relative to the lightest feasible design is unbounded---a genuine decision failure. All five main $25{,}000$-candidate pools contain feasible candidates (Table~\ref{tab:eval_pools}), so $\infty$ in a main-protocol table always denotes a model failure; N/A arises only in the revision protocol, where re-sampling around the feasibility frontier can empty the feasible set for a card. Reference baselines are random ranking, lightest-first ranking, a pseudo-bootstrap design score and an oracle upper bound that uses oracle feasibility directly to verify metric well-formedness. The pseudo-bootstrap score is a non-neural sanity-check rule used only to verify that the evaluation pipeline separates curve reconstruction from feasible-candidate retrieval; it is not used to support any claim about surrogate superiority.

\subsection*{Cross-material transfer}
For cross-material transfer, each per-material checkpoint is evaluated on every other material's held-out pool. Only the dimensionless scaling and material vector $m^*$ are swapped to those of the target card; no retraining or fine-tuning is performed. The transfer matrix thus isolates the inductive bias the network absorbed during training from the dimensional-scale change that the Buckingham-$\pi$ formulation neutralises. All transfer numbers in the manuscript are reported under PINN-energy unless noted; the supplementary material includes the equivalent matrices for MLP-softplus and PINN-full.

\subsection*{Reporting and reproducibility}
The manuscript reports model architecture, input representation, training data generation, random seeds, optimisation settings, checkpoint metadata, evaluation metrics and baseline comparisons to support reproducibility. The candidate pools, oracle labels, dimensionless inputs, training hyperparameters, optimiser state, checkpoint hashes, evaluation scripts and machine-readable metric tables are bundled in the repository together with this manuscript. The supplementary material provides full per-method, per-material and per-transfer tables that underline the compact summaries in the main text.

\section*{Data availability}
The numerical benchmark archive, including candidate pools, generated metrics, figures, configuration files and run metadata, is available in the project repository at \repositoryurl{}.

\section*{Code availability}
The code is available at \repositoryurl{}. The repository contains scripts for candidate generation, oracle evaluation, model training, metric aggregation and figure reproduction.

\section*{Acknowledgements}
The authors received no specific funding for this work.

\section*{Author contributions}
D.C. conceived the benchmark framing, implemented and curated the numerical workflow, analysed the results, prepared the figures and tables, and wrote the manuscript. A.C. supervised the research direction, provided conceptual guidance, and reviewed the manuscript.

\section*{Competing interests}
The authors declare no competing interests.

\section*{Use of generative AI in manuscript preparation}
Generative AI assistants were used for language polishing and to suggest restructuring of paragraphs in the introduction and discussion. All scientific content, methodology, code, oracle definitions, surrogate training, evaluation procedures, numerical results and interpretation are the authors' own. No generative AI was used to produce results, data, figures or citations; all citations were verified against the primary sources before inclusion.

\end{document}


\maketitle

\section*{Supplementary material and audit trail}

The supplementary material provides an audit trail for the numerical benchmark. It contains the aggregate metric files, candidate-level curve-error summaries, supplementary \LaTeX{} tables, and diagnostic figures used to reproduce and verify the reported results. These materials extend the declared-oracle benchmark reported in the main text; they should not be interpreted as an independent experimental or finite-element validation dataset.

\subsection*{Supplementary tables}

This section is the table-level audit trail for the compact results reported in the main manuscript. All entries are computed from the declared-oracle evaluation pipeline on the same fixed held-out candidate pools used in the main protocol: $25{,}000$ candidates per material card. The tables therefore expand the manuscript results; they do not introduce an additional experimental or finite-element validation protocol.

The tables are organized from sanity checks to the full transfer audit. Table S1 verifies the metric pipeline using non-neural reference rules and the oracle upper bound. Table S2 reports the complete method-level benchmark for material-specific, pooled, and reference models. Table S3 gives the strongest observed cross-material transfer cases for each target card. Table S4 provides the full source--target transfer matrix and is the numerical source for the transfer heatmaps in the diagnostic figures.

\paragraph*{Source of the tabulated values.}
The values are exported from the benchmark repository after evaluating trained surrogates against the declared oracle. For each candidate, the oracle supplies feasibility and mass, while each surrogate supplies a predicted force--displacement response and a predicted design ranking. The tabulated values are then obtained by grouping candidates by material card, model family, and, for transfer analyses, source--target material pair. No supplementary-table values are manually read from figures.

\paragraph*{Reporting conventions.}
\emph{nRMSE} measures normalized force--displacement curve error; lower is better. \emph{Violation} is the fraction of surrogate-selected feasible candidates rejected by the oracle; lower is safer. \emph{P@10} is the fraction of the ten highest-ranked surrogate candidates that are oracle-feasible; higher is better. \emph{Regret@10} is the mass gap, in grams, between the best feasible candidate found in the surrogate top ten and the lightest oracle-feasible candidate in the same pool; lower is better. Thus, regret is a decision metric rather than a curve-fitting metric.

Two non-finite entries have different meanings. \textbf{N/A} denotes a material pool with no oracle-feasible candidate under the fixed fixture, making retrieval and regret undefined. \textbf{$\infty$} denotes a non-empty feasible pool where the model retrieves no oracle-feasible candidate in its top ten. This is counted as a decision failure, not as missing data. In the five main held-out pools, $\infty$ may appear, while N/A should not appear.

The boxed table frames are only visual separators between audit blocks. They do not denote statistical significance.

\Needspace{8\baselineskip}
\subsubsection*{Supplementary Table S1: baseline and oracle sanity checks}

Table S1 reports non-neural reference rules and the oracle upper bound. It checks that the metric pipeline behaves as expected before learned surrogates are interpreted. The oracle row verifies the achievable top-ranked feasible retrieval under the declared fixture. Random, lightest-first, and pseudo-bootstrap references show what performance is obtainable without a trained material-conditioned surrogate. The pseudo-bootstrap baseline is included only as a pipeline sanity check and is not interpreted as an optimized design method.

\begin{table}[H]
\centering
\small
\caption{\textbf{Non-neural baselines and the oracle upper bound (main protocol).} Reference rankers evaluated on the same $25{,}000$-candidate held-out pools as the learned surrogates. The oracle upper bound uses oracle feasibility directly and confirms that the ranking metrics are well formed (P@10 $=1.000$, regret $=0$). The lightest-first and pseudo-bootstrap rankers ignore feasibility, so their violation rates are high and they retrieve no feasible design on the stiffer cards. Lower nRMSE, violation and regret are better; higher P@10 is better.}
\label{tab:s_baselines}

\resizebox{\textwidth}{!}{%
\begin{tabular}{llrrrr}
\toprule
Material & Baseline & nRMSE $\downarrow$ & Violation $\downarrow$ & P@10 $\uparrow$ & Regret@10 [g] $\downarrow$ \\
\midrule
PA12 & Oracle upper bound & 0.000 & 0.000 & 1.000 & 0.00 \\
PA12 & Random & 0.421 & 0.825 & 0.400 & 9.64 \\
PA12 & Lightest & 0.421 & 0.825 & 1.000 & 0.00 \\
PA12 & Pseudo-bootstrap & 0.421 & 0.825 & 1.000 & 0.00 \\
\addlinespace

PLA & Oracle upper bound & 0.000 & 0.000 & 1.000 & 0.00 \\
PLA & Random & 0.327 & 0.750 & 0.200 & 19.57 \\
PLA & Lightest & 0.327 & 0.750 & 0.000 & $\infty$ \\
PLA & Pseudo-bootstrap & 0.327 & 0.750 & 0.000 & $\infty$ \\
\addlinespace

PETG & Oracle upper bound & 0.000 & 0.000 & 1.000 & 0.00 \\
PETG & Random & 0.321 & 0.825 & 0.100 & 21.52 \\
PETG & Lightest & 0.321 & 0.825 & 0.000 & $\infty$ \\
PETG & Pseudo-bootstrap & 0.321 & 0.825 & 0.000 & $\infty$ \\
\addlinespace

TPU & Oracle upper bound & 0.000 & 0.000 & 1.000 & 0.00 \\
TPU & Random & 1.138 & 0.354 & 0.500 & 13.57 \\
TPU & Lightest & 1.138 & 0.354 & 0.000 & $\infty$ \\
TPU & Pseudo-bootstrap & 1.138 & 0.354 & 0.000 & $\infty$ \\
\addlinespace

PA-CF & Oracle upper bound & 0.000 & 0.000 & 1.000 & 0.00 \\
PA-CF & Random & 0.339 & 0.872 & 0.200 & 13.81 \\
PA-CF & Lightest & 0.339 & 0.872 & 0.100 & 0.00 \\
PA-CF & Pseudo-bootstrap & 0.339 & 0.872 & 0.100 & 0.00 \\
\bottomrule
\end{tabular}%
}

\par\vspace{3pt}
\begin{minipage}{\textwidth}
\footnotesize\raggedright
\textit{Columns.} nRMSE: curve fidelity (lower is better); Violation: fraction of predicted-feasible candidates the oracle rejects; P@10: top-10 feasible-retrieval precision; Regret@10: mass gap in grams to the lightest oracle-feasible design. 
\textit{Symbols.} $\infty$ marks a non-empty feasible pool from which the model retrieved no feasible candidate in its top ten (a decision failure, not a missing pool); N/A would mark a pool with no oracle-feasible candidate, which does not occur in the main $25{,}000$-candidate pools. All rows use the main protocol.
\end{minipage}
\end{table}

\Needspace{8\baselineskip}
\subsubsection*{Supplementary Table S2: complete method-level benchmark}

Table S2 expands the compact method comparison from the manuscript. It groups baseline rules, material-specific models, and pooled material-conditioned models under the same metric definitions. Its purpose is to show where curve reconstruction and design usefulness diverge: a model can obtain acceptable nRMSE while still producing high violation, weak P@10, or infinite regret if its top-ranked designs are not oracle-feasible.

\begin{table}[H]
\centering
\scriptsize
\caption{\textbf{Complete per-card method metrics (main protocol).} All surrogate scopes and reference baselines on the $25{,}000$-candidate held-out pools, grouped by material card and then by scope (per-material specialist, pooled material-conditioned model, non-neural reference). This table is the full version of the compact summaries in the main text. Lower nRMSE, violation and regret are better; higher P@10 is better.}
\label{tab:s_main_method_metrics_full}

\resizebox{\textwidth}{!}{%
\begin{tabular}{lllrrrr}
\toprule
Material & Scope & Method & nRMSE $\downarrow$ & Violation $\downarrow$ & P@10 $\uparrow$ & Regret@10 [g] $\downarrow$ \\
\midrule
PA12 & Per-material & MLP-softplus & 0.596 & 0.000 & 1.000 & 1.03 \\
PA12 & Per-material & PINN-energy & 0.184 & 0.000 & 1.000 & 0.28 \\
PA12 & Per-material & PINN-full & 0.704 & 0.000 & 1.000 & 0.57 \\
\addlinespace
PA12 & Pooled & MLP-softplus & 0.597 & 0.000 & 1.000 & 0.00 \\
PA12 & Pooled & PINN-energy & 0.551 & 0.000 & 0.900 & 1.68 \\
PA12 & Pooled & PINN-full & 0.594 & 0.000 & 0.500 & 16.07 \\
\addlinespace
PA12 & Reference & Oracle upper bound & 0.000 & 0.000 & 1.000 & 0.00 \\
PA12 & Reference & Random & 0.421 & 0.825 & 0.400 & 9.64 \\
PA12 & Reference & Lightest & 0.421 & 0.825 & 1.000 & 0.00 \\
PA12 & Reference & Pseudo-bootstrap & 0.421 & 0.825 & 1.000 & 0.00 \\
\midrule

PLA & Per-material & MLP-softplus & 0.462 & 0.000 & 0.500 & 5.82 \\
PLA & Per-material & PINN-energy & 0.119 & 0.000 & 0.200 & 5.00 \\
PLA & Per-material & PINN-full & 0.408 & 0.000 & 0.000 & $\infty$ \\
\addlinespace
PLA & Pooled & MLP-softplus & 0.408 & 0.000 & 0.000 & $\infty$ \\
PLA & Pooled & PINN-energy & 0.386 & 0.000 & 0.700 & 8.31 \\
PLA & Pooled & PINN-full & 0.407 & 0.000 & 0.000 & $\infty$ \\
\addlinespace
PLA & Reference & Oracle upper bound & 0.000 & 0.000 & 1.000 & 0.00 \\
PLA & Reference & Random & 0.327 & 0.750 & 0.200 & 19.57 \\
PLA & Reference & Lightest & 0.327 & 0.750 & 0.000 & $\infty$ \\
PLA & Reference & Pseudo-bootstrap & 0.327 & 0.750 & 0.000 & $\infty$ \\
\midrule

PETG & Per-material & MLP-softplus & 0.408 & 0.000 & 0.000 & $\infty$ \\
PETG & Per-material & PINN-energy & 0.405 & 0.000 & 1.000 & 5.92 \\
PETG & Per-material & PINN-full & 0.408 & 0.000 & 0.700 & 5.92 \\
\addlinespace
PETG & Pooled & MLP-softplus & 0.409 & 0.000 & 0.000 & $\infty$ \\
PETG & Pooled & PINN-energy & 0.388 & 0.000 & 0.000 & $\infty$ \\
PETG & Pooled & PINN-full & 0.408 & 1.000 & 0.100 & 26.07 \\
\addlinespace
PETG & Reference & Oracle upper bound & 0.000 & 0.000 & 1.000 & 0.00 \\
PETG & Reference & Random & 0.321 & 0.825 & 0.100 & 21.52 \\
PETG & Reference & Lightest & 0.321 & 0.825 & 0.000 & $\infty$ \\
PETG & Reference & Pseudo-bootstrap & 0.321 & 0.825 & 0.000 & $\infty$ \\
\midrule

TPU & Per-material & MLP-softplus & 0.670 & 0.000 & 0.000 & $\infty$ \\
TPU & Per-material & PINN-energy & 0.252 & 0.104 & 0.000 & $\infty$ \\
TPU & Per-material & PINN-full & 0.694 & 0.000 & 0.000 & $\infty$ \\
\addlinespace
TPU & Pooled & MLP-softplus & 0.694 & 0.000 & 0.000 & $\infty$ \\
TPU & Pooled & PINN-energy & 0.623 & 0.000 & 0.000 & $\infty$ \\
TPU & Pooled & PINN-full & 0.690 & 0.000 & 0.900 & 12.71 \\
\addlinespace
TPU & Reference & Oracle upper bound & 0.000 & 0.000 & 1.000 & 0.00 \\
TPU & Reference & Random & 1.138 & 0.354 & 0.500 & 13.57 \\
TPU & Reference & Lightest & 1.138 & 0.354 & 0.000 & $\infty$ \\
TPU & Reference & Pseudo-bootstrap & 1.138 & 0.354 & 0.000 & $\infty$ \\
\midrule

PA-CF & Per-material & MLP-softplus & 0.260 & 0.399 & 0.200 & 0.00 \\
PA-CF & Per-material & PINN-energy & 0.402 & 0.000 & 1.000 & 6.89 \\
PA-CF & Per-material & PINN-full & 0.410 & 0.000 & 0.100 & 0.00 \\
\addlinespace
PA-CF & Pooled & MLP-softplus & 0.410 & 0.000 & 0.200 & 0.00 \\
PA-CF & Pooled & PINN-energy & 0.383 & 1.000 & 0.000 & $\infty$ \\
PA-CF & Pooled & PINN-full & 0.409 & 0.000 & 0.200 & 14.83 \\
\addlinespace
PA-CF & Reference & Oracle upper bound & 0.000 & 0.000 & 1.000 & 0.00 \\
PA-CF & Reference & Random & 0.339 & 0.872 & 0.200 & 13.81 \\
PA-CF & Reference & Lightest & 0.339 & 0.872 & 0.100 & 0.00 \\
PA-CF & Reference & Pseudo-bootstrap & 0.339 & 0.872 & 0.100 & 0.00 \\
\bottomrule
\end{tabular}%
}

\par\vspace{3pt}
\begin{minipage}{\textwidth}
\footnotesize\raggedright
\textit{Columns.} nRMSE: curve fidelity (lower is better); Violation: fraction of predicted-feasible candidates the oracle rejects; P@10: top-10 feasible-retrieval precision; Regret@10: mass gap in grams to the lightest oracle-feasible design. 
\textit{Symbols.} $\infty$ marks a non-empty feasible pool from which the model retrieved no feasible candidate in its top ten (a decision failure, not a missing pool); N/A would mark a pool with no oracle-feasible candidate, which does not occur in the main $25{,}000$-candidate pools. All rows use the main protocol.
\end{minipage}
\end{table}

\Needspace{8\baselineskip}
\subsubsection*{Supplementary Table S3: strongest transfer case for each target}

Table S3 reduces the full source--target transfer matrix to the best observed cross-material cases for each target material. The donor material is not assumed in advance; it is selected from the evaluated source cards and surrogate variants. The table is therefore a readable summary of asymmetric transfer behaviour rather than evidence of universal material-invariant learning.

\begin{table}[H]
\centering
\scriptsize
\caption{\textbf{Strongest cross-material transfer by target card (main protocol).} For each evaluation (target) card, the five source checkpoints with the lowest curve error, ordered by nRMSE. The complete matrix is given in Table~\ref{tab:s_transfer_full}. Low curve error on a target does not imply feasible retrieval: every transfer onto PETG, PLA and TPU fails to retrieve a feasible design in its top ten ($\infty$), even at sub-$0.3$ nRMSE.}
\label{tab:s_best_transfer_by_target}

\resizebox{\textwidth}{!}{%
\begin{tabular}{lllrrrr}
\toprule
Train & Eval & Method & nRMSE $\downarrow$ & Violation $\downarrow$ & P@10 $\uparrow$ & Regret@10 [g] $\downarrow$ \\
\midrule
TPU & PA12 & PINN-energy & 0.321 & 0.000 & 1.000 & 0.00 \\
PA-CF & PA12 & PINN-energy & 0.584 & 0.000 & 1.000 & 0.00 \\
TPU & PA12 & MLP-softplus & 0.585 & 0.000 & 1.000 & 0.00 \\
PETG & PA12 & PINN-energy & 0.589 & 0.000 & 1.000 & 0.00 \\
PETG & PA12 & PINN-full & 0.595 & 0.000 & 1.000 & 0.00 \\
\addlinespace

PA12 & PLA & PINN-energy & 0.267 & 0.000 & 0.000 & $\infty$ \\
TPU & PLA & PINN-energy & 0.271 & 0.000 & 0.000 & $\infty$ \\
PA-CF & PLA & PINN-energy & 0.402 & 0.000 & 0.000 & $\infty$ \\
TPU & PLA & MLP-softplus & 0.402 & 0.000 & 0.000 & $\infty$ \\
PETG & PLA & PINN-energy & 0.404 & 0.000 & 0.000 & $\infty$ \\
\addlinespace

TPU & PETG & PINN-energy & 0.266 & 0.000 & 0.000 & $\infty$ \\
PA-CF & PETG & PINN-energy & 0.403 & 0.000 & 0.000 & $\infty$ \\
TPU & PETG & MLP-softplus & 0.403 & 0.000 & 0.000 & $\infty$ \\
PA12 & PETG & MLP-softplus & 0.408 & 0.000 & 0.000 & $\infty$ \\
PA-CF & PETG & PINN-full & 0.408 & 0.000 & 0.000 & $\infty$ \\
\addlinespace

PA-CF & TPU & PINN-energy & 0.673 & 0.000 & 0.000 & $\infty$ \\
PETG & TPU & PINN-energy & 0.681 & 0.000 & 0.000 & $\infty$ \\
PETG & TPU & PINN-full & 0.691 & 0.000 & 0.000 & $\infty$ \\
PA12 & TPU & MLP-softplus & 0.692 & 0.000 & 0.000 & $\infty$ \\
PETG & TPU & MLP-softplus & 0.693 & 0.000 & 0.000 & $\infty$ \\
\addlinespace

PA12 & PA-CF & PINN-energy & 0.297 & 0.000 & 0.100 & 0.00 \\
TPU & PA-CF & PINN-energy & 0.303 & 0.000 & 0.100 & 0.00 \\
PLA & PA-CF & PINN-energy & 0.309 & 0.872 & 0.100 & 0.00 \\
TPU & PA-CF & MLP-softplus & 0.403 & 0.000 & 0.100 & 0.00 \\
PETG & PA-CF & PINN-energy & 0.405 & 0.000 & 0.100 & 0.00 \\
\bottomrule
\end{tabular}%
}

\par\vspace{3pt}
\begin{minipage}{\textwidth}
\footnotesize\raggedright
\textit{Columns.} Train/Eval: source card the checkpoint was trained on and target card it is evaluated on; only the dimensionless scaling and material vector are swapped, with no retraining or fine-tuning. nRMSE: curve fidelity; Violation: fraction of predicted-feasible candidates the oracle rejects; P@10: top-10 feasible-retrieval precision; Regret@10: mass gap in grams. $\infty$ marks a non-empty target pool from which the transferred model retrieved no feasible candidate in its top ten. All rows use the main protocol.
\end{minipage}
\end{table}

\Needspace{8\baselineskip}
\subsubsection*{Supplementary Table S4: complete cross-material transfer matrix}

Table S4 reports all evaluated source--target material pairs and surrogate variants. Rows are grouped by target card so that deployment failures can be inspected from the perspective of a new material. This is the most detailed supplementary table and the numerical source for the transfer heatmaps in Figures S2--S4.

\begin{longtable}{@{}lllrrrr@{}}
\caption{\textbf{Full cross-material transfer matrix.} Complete source--target transfer results for the main protocol. Rows are grouped by target material and ordered by nRMSE.}
\label{tab:s_transfer_full}\\

\toprule
Train & Eval & Method & nRMSE $\downarrow$ & Viol. $\downarrow$ & P@10 $\uparrow$ & Regret [g] $\downarrow$ \\
\midrule
\endfirsthead

\caption[]{Full cross-material transfer matrix (continued).}\\
\toprule
Train & Eval & Method & nRMSE $\downarrow$ & Viol. $\downarrow$ & P@10 $\uparrow$ & Regret [g] $\downarrow$ \\
\midrule
\endhead

\midrule
\multicolumn{7}{r}{\scriptsize Continued on next page}\\
\endfoot

\bottomrule
\multicolumn{7}{@{}p{\dimexpr\textwidth\relax}@{}}{\scriptsize
Train/Eval denote source and target material cards. Transfer is evaluated by swapping only scaling and material vector, without fine-tuning. $\infty$ denotes a non-empty target pool where no feasible design was retrieved in the top ten.
}\\
\endlastfoot

TPU & PA12 & PINN-energy & 0.321 & 0.000 & 1.000 & 0.00 \\
PA-CF & PA12 & PINN-energy & 0.584 & 0.000 & 1.000 & 0.00 \\
TPU & PA12 & MLP-softplus & 0.585 & 0.000 & 1.000 & 0.00 \\
PETG & PA12 & PINN-energy & 0.589 & 0.000 & 1.000 & 0.00 \\
PETG & PA12 & PINN-full & 0.595 & 0.000 & 1.000 & 0.00 \\
PETG & PA12 & MLP-softplus & 0.596 & 0.000 & 1.000 & 0.00 \\
PA-CF & PA12 & PINN-full & 0.596 & 0.000 & 1.000 & 0.00 \\
TPU & PA12 & PINN-full & 0.597 & 0.000 & 1.000 & 0.00 \\
PLA & PA12 & PINN-full & 0.597 & 0.000 & 1.000 & 0.00 \\
PLA & PA12 & PINN-energy & 0.683 & 0.000 & 1.000 & 0.00 \\
PA-CF & PA12 & MLP-softplus & 0.715 & 0.000 & 1.000 & 0.00 \\
PLA & PA12 & MLP-softplus & 0.736 & 0.000 & 1.000 & 0.00 \\

\midrule
PA12 & PLA & PINN-energy & 0.267 & 0.000 & 0.000 & $\infty$ \\
TPU & PLA & PINN-energy & 0.271 & 0.000 & 0.000 & $\infty$ \\
PA-CF & PLA & PINN-energy & 0.402 & 0.000 & 0.000 & $\infty$ \\
TPU & PLA & MLP-softplus & 0.402 & 0.000 & 0.000 & $\infty$ \\
PETG & PLA & PINN-energy & 0.404 & 0.000 & 0.000 & $\infty$ \\
PETG & PLA & PINN-full & 0.407 & 0.000 & 0.000 & $\infty$ \\
PA12 & PLA & MLP-softplus & 0.408 & 0.000 & 0.000 & $\infty$ \\
PETG & PLA & MLP-softplus & 0.408 & 0.000 & 0.000 & $\infty$ \\
PA-CF & PLA & PINN-full & 0.408 & 0.000 & 0.000 & $\infty$ \\
TPU & PLA & PINN-full & 0.408 & 0.000 & 0.000 & $\infty$ \\
PA-CF & PLA & MLP-softplus & 0.482 & 0.000 & 0.000 & $\infty$ \\
PA12 & PLA & PINN-full & 0.501 & 0.000 & 0.000 & $\infty$ \\

\midrule
TPU & PETG & PINN-energy & 0.266 & 0.000 & 0.000 & $\infty$ \\
PA-CF & PETG & PINN-energy & 0.403 & 0.000 & 0.000 & $\infty$ \\
TPU & PETG & MLP-softplus & 0.403 & 0.000 & 0.000 & $\infty$ \\
PA12 & PETG & MLP-softplus & 0.408 & 0.000 & 0.000 & $\infty$ \\
PA-CF & PETG & PINN-full & 0.408 & 0.000 & 0.000 & $\infty$ \\
TPU & PETG & PINN-full & 0.409 & 0.000 & 0.000 & $\infty$ \\
PLA & PETG & PINN-full & 0.409 & 0.000 & 0.000 & $\infty$ \\
PLA & PETG & MLP-softplus & 0.425 & 0.000 & 0.000 & $\infty$ \\
PA12 & PETG & PINN-full & 0.438 & 0.000 & 0.000 & $\infty$ \\
PA-CF & PETG & MLP-softplus & 0.440 & 0.000 & 0.000 & $\infty$ \\
PA12 & PETG & PINN-energy & 0.597 & 0.000 & 0.000 & $\infty$ \\
PLA & PETG & PINN-energy & 1.287 & 0.000 & 0.000 & $\infty$ \\

\midrule
PA-CF & TPU & PINN-energy & 0.673 & 0.000 & 0.000 & $\infty$ \\
PETG & TPU & PINN-energy & 0.681 & 0.000 & 0.000 & $\infty$ \\
PETG & TPU & PINN-full & 0.691 & 0.000 & 0.000 & $\infty$ \\
PA12 & TPU & MLP-softplus & 0.692 & 0.000 & 0.000 & $\infty$ \\
PETG & TPU & MLP-softplus & 0.693 & 0.000 & 0.000 & $\infty$ \\
PA-CF & TPU & PINN-full & 0.693 & 0.000 & 0.000 & $\infty$ \\
PLA & TPU & PINN-full & 0.694 & 0.000 & 0.000 & $\infty$ \\
PLA & TPU & MLP-softplus & 1.448 & 0.354 & 0.000 & $\infty$ \\
PA-CF & TPU & MLP-softplus & 1.453 & 0.354 & 0.000 & $\infty$ \\
PA12 & TPU & PINN-full & 1.506 & 0.354 & 0.000 & $\infty$ \\
PLA & TPU & PINN-energy & 1.909 & 0.354 & 0.000 & $\infty$ \\
PA12 & TPU & PINN-energy & 11.342 & 0.000 & 0.000 & $\infty$ \\

\midrule
PA12 & PA-CF & PINN-energy & 0.297 & 0.000 & 0.100 & 0.00 \\
TPU & PA-CF & PINN-energy & 0.303 & 0.000 & 0.100 & 0.00 \\
PLA & PA-CF & PINN-energy & 0.309 & 0.872 & 0.100 & 0.00 \\
TPU & PA-CF & MLP-softplus & 0.403 & 0.000 & 0.100 & 0.00 \\
PETG & PA-CF & PINN-energy & 0.405 & 0.000 & 0.100 & 0.00 \\
PETG & PA-CF & PINN-full & 0.409 & 0.000 & 0.100 & 0.00 \\
PA12 & PA-CF & MLP-softplus & 0.410 & 0.000 & 0.100 & 0.00 \\
PETG & PA-CF & MLP-softplus & 0.410 & 0.000 & 0.100 & 0.00 \\
TPU & PA-CF & PINN-full & 0.410 & 0.000 & 0.100 & 0.00 \\
PLA & PA-CF & PINN-full & 0.410 & 0.000 & 0.100 & 0.00 \\
PLA & PA-CF & MLP-softplus & 0.616 & 0.000 & 0.100 & 0.00 \\
PA12 & PA-CF & PINN-full & 0.638 & 0.000 & 0.100 & 0.00 \\

\end{longtable}

\subsection*{Supplementary diagnostic figures}

The diagnostic figures visualize the same audit logic as the tables. Figure S1 checks baseline behaviour on held-out candidates. Figures S2--S4 show cross-material transfer from three complementary perspectives: curve reconstruction, feasible-candidate retrieval, and constraint-safety behaviour. These views are intentionally separated because a low aggregate curve-error score can still coincide with poor top-ranked feasible retrieval or unsafe selection.

\begin{figure}[!htbp]
\centering
\includegraphics[width=\linewidth]{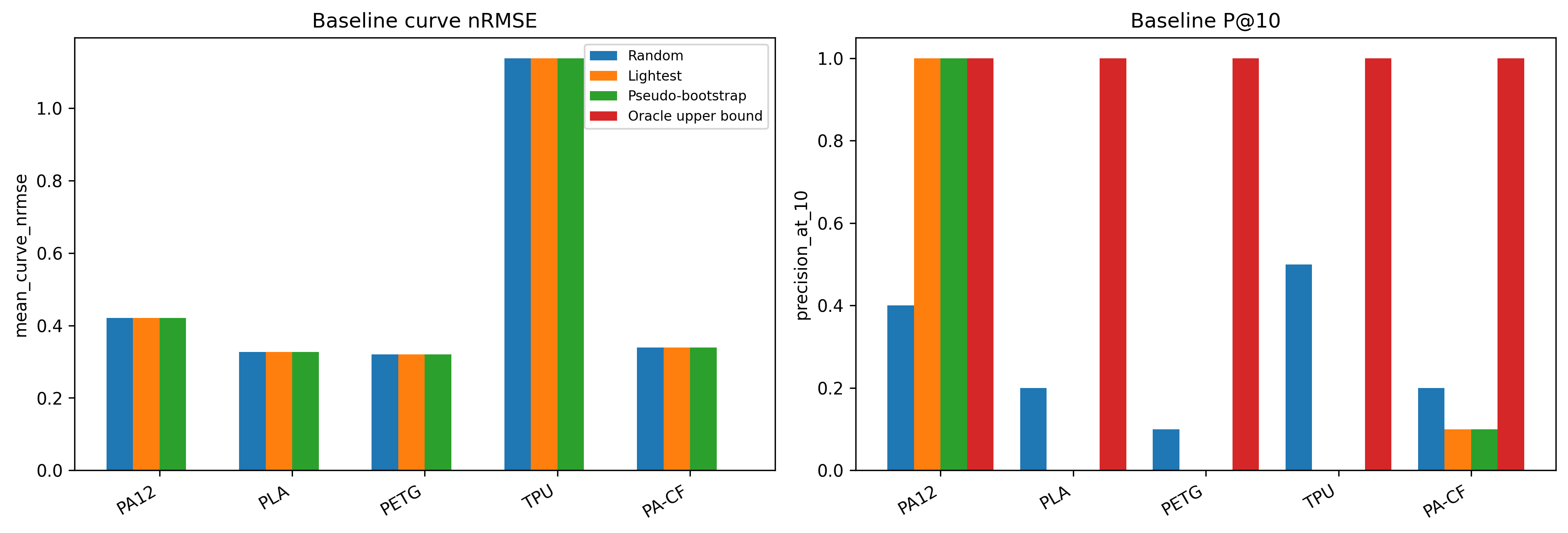}
\caption{\textbf{Baseline behaviour on held-out candidates.}
The oracle upper bound verifies that the ranking metrics are well formed under the declared oracle. Random, lightest-first, and pseudo-bootstrap baselines provide non-neural references for curve error and top-10 retrieval. These baselines are not intended as optimized design methods; they define sanity checks for the learned surrogates and help separate curve reconstruction from feasible-candidate retrieval.}
\label{fig:s_baselines}
\end{figure}

\begin{figure}[!htbp]
\centering
\includegraphics[width=\linewidth]{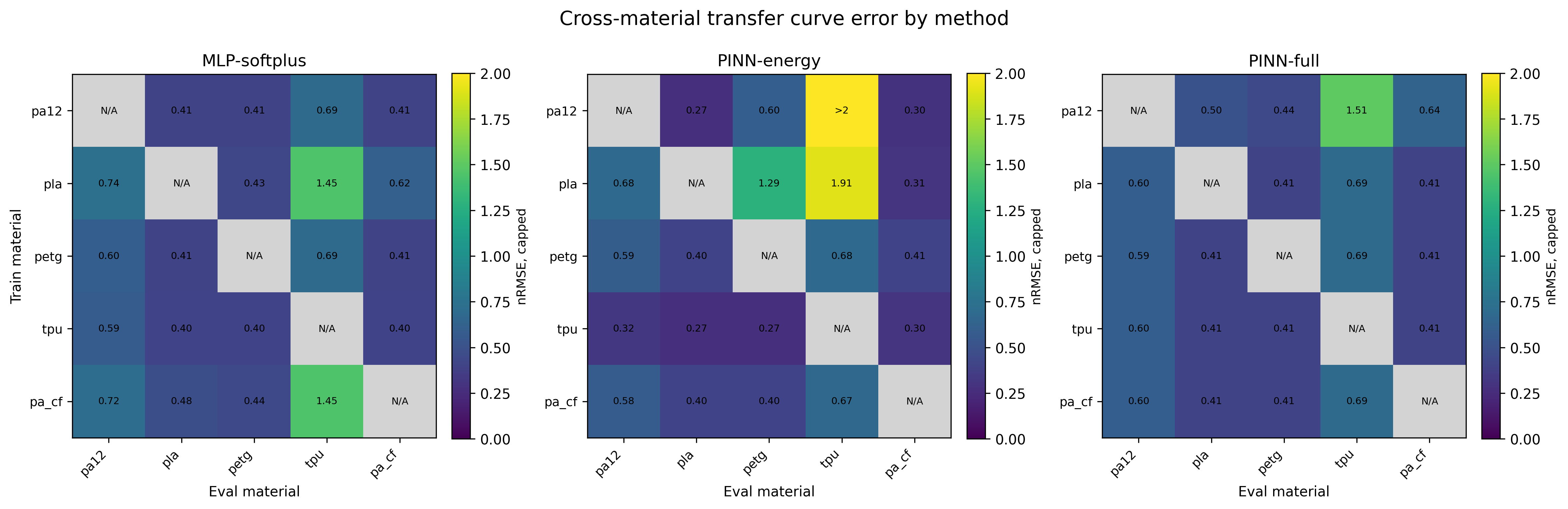}
\caption{\textbf{Cross-material transfer curve error for all surrogate variants.}
Each panel reports nRMSE for one model family evaluated under cross-material transfer. Rows denote the material card used for training and columns denote the material card used for evaluation. Diagonal entries correspond to omitted self-material cases, which are reported separately in the material-specific evaluation. nRMSE values above 2.0 are saturated for readability; the underlying machine-readable files contain the unsaturated values.}
\label{fig:s_transfer_all_nrmse}
\end{figure}

\begin{figure}[!htbp]
\centering
\includegraphics[width=\linewidth]{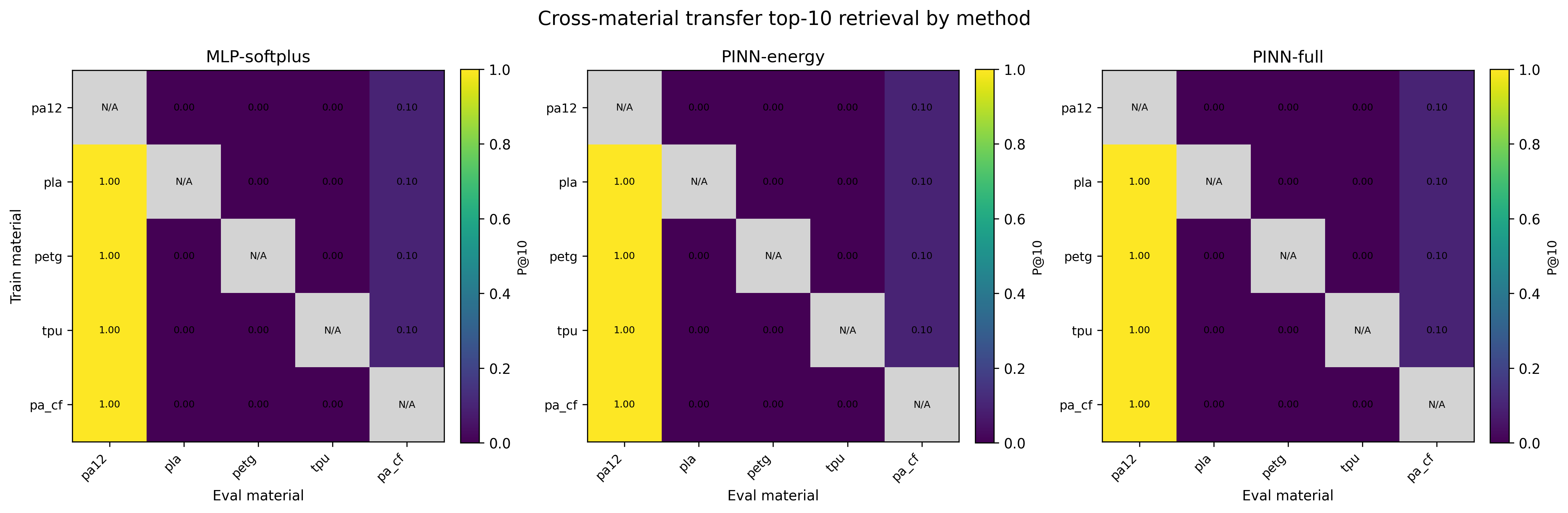}
\caption{\textbf{Cross-material transfer top-10 retrieval for all surrogate variants.}
Each panel reports precision@10 for oracle-feasible candidates under the predicted design score. The figure shows that moderate curve fidelity does not necessarily imply successful top-ranked feasible design retrieval, especially for non-PA12 target materials.}
\label{fig:s_transfer_all_p10}
\end{figure}

\begin{figure}[!htbp]
\centering
\includegraphics[width=\linewidth]{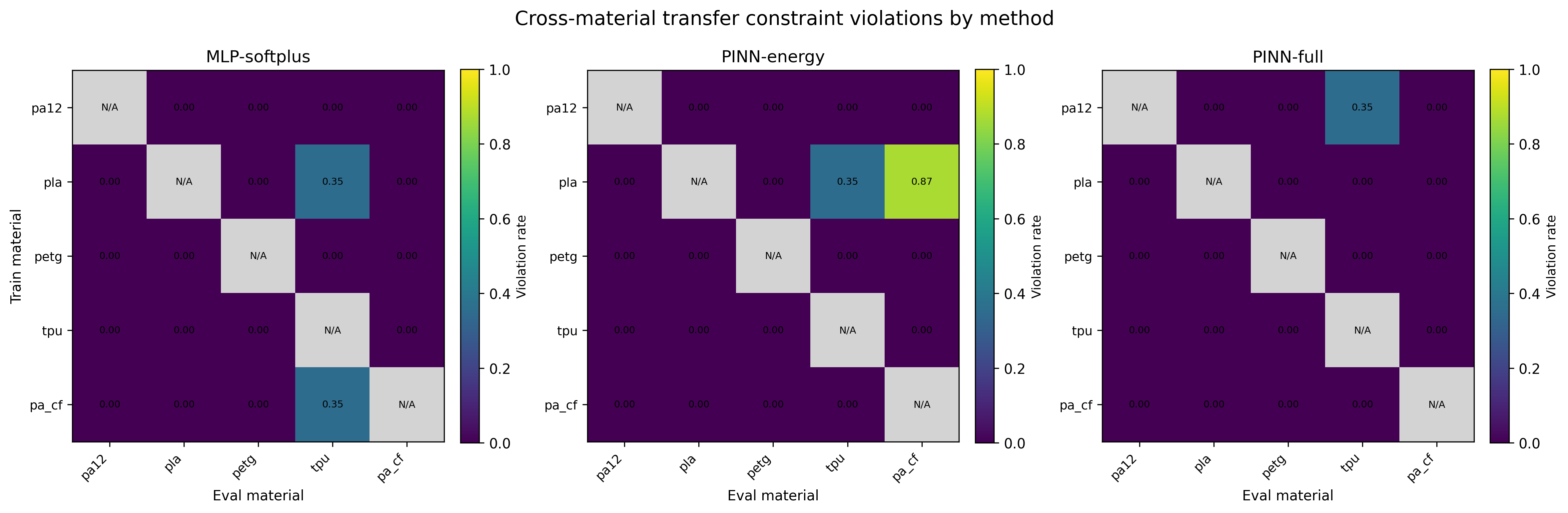}
\caption{\textbf{Cross-material transfer constraint violations for all surrogate variants.}
Each panel reports the fraction of predicted-feasible candidates that violate oracle feasibility constraints. The figure diagnoses unsafe selection behaviour under material-regime mismatch. Low curve error and low violation rate should therefore be interpreted as complementary, not interchangeable, properties of a surrogate.}
\label{fig:s_transfer_all_violation}
\end{figure}